\documentclass[conference]{IEEEtran}
\IEEEoverridecommandlockouts
\usepackage{cite}
\usepackage{amsmath,amssymb,amsfonts}
\usepackage{algorithmic}
\usepackage{graphicx}
\usepackage{textcomp}
\usepackage{xcolor}
\usepackage{color}
\usepackage{booktabs}
\usepackage{subfigure}
\usepackage{float}
\usepackage{makecell}
\usepackage{multirow}
\usepackage{bbding}

\def\BibTeX{{\rm B\kern-.05em{\sc i\kern-.025em b}\kern-.08em
    T\kern-.1667em\lower.7ex\hbox{E}\kern-.125emX}}
\begin{document}

\title{Attribute analysis with synthetic dataset for person re-identification \\
}


\author{\IEEEauthorblockN{Suncheng Xiang$^{1}$, Yuzhuo Fu$^{1}$, Guanjie You$^{1}$, Ting Liu$^{1}$}
\IEEEauthorblockA{\textit{$^{1}$School of Electronic Information and Electrical Engineering} \\
\textit{Shanghai Jiao Tong University}\\
Shanghai, China \\
\{xiangsuncheng17, yzfu, ygj.sjtu, louisa\_liu\}@sjtu.edu.cn}
}

\maketitle

\begin{abstract}
Person re-identification (re-ID) plays an important role in applications such as public security and video surveillance. Recently, learning from synthetic data, which benefits from the popularity of synthetic data engine, have achieved remarkable performance. However, existing synthetic datasets are in small size and lack of diversity, which hinders the development of person re-ID in real-world scenarios. To address this problem, firstly, we develop a large-scale synthetic data engine, the salient characteristic of this engine is controllable. Based on it, we build a large-scale synthetic dataset, which are diversified and customized from different attributes, such as illumination and viewpoint. Secondly, we quantitatively analyze the influence of dataset attributes on re-ID system. To our best knowledge, this is the first attempt to explicitly dissect person re-ID from the aspect of attribute on synthetic dataset. Comprehensive experiments help us have a deeper understanding of the fundamental problems in person re-ID. Our research also provides useful insights for dataset building and future practical usage.

\end{abstract}


\section{Introduction}
\label{sec1}
Person re-ID aims to identify images of the same person from large number of cameras views in different places, which has attracted lots of interests and attention in both academia and industry. Encouraged by the remarkable success of deep learning methods~\cite{li2014deepreid,he2016deep} and the availability of re-ID datasets~\cite{zheng2015scalable,ristani2016performance}, performance of person re-ID has been significantly boosted. For example, the the rank-1 accuracy of single query on Market-1501~\cite{zheng2015scalable} has been improved from 43.8\%~\cite{liao2015person} to 91.2\%~\cite{kalayeh2018human}, the rank-1 accuracy on DukeMTMC-reID~\cite{ristani2016performance} has been improved from 25.13\%~\cite{zheng2015scalable} to 85.95\%~\cite{kalayeh2018human}. Currently, these performance gains comes only when a large diversity of training data is available, which is at the price of large amount of accurate annotations obtained by expensive human labor. Accordingly, real applications have to cope with challenges like complex lighting and scene variations, which current real datasets might fail to address~\cite{xiang2020unsupervisedperson}.

\begin{figure}
\centerline{\includegraphics[width=\linewidth]{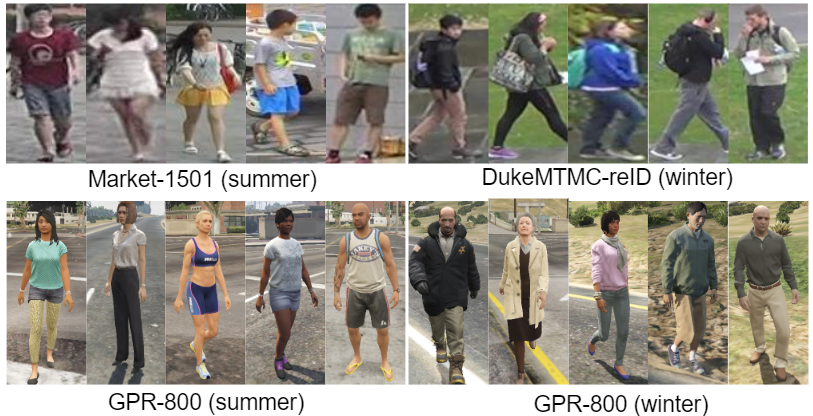}}
\caption{Illustration of the examples between Market-1501 (upper-left) and DukeMTMC-reID (upper-right). It can be obviously found that there exists difference among different datasets in light, background or weather. However, our proposed large-scale dataset GPR-800 (bottom) always has large variances, high resolution and different backgrounds.}
\label{fig1}
\end{figure}

To address this issue, many successful person re-ID approaches~\cite{barbosa2018looking,bak2018domain,sun2019dissecting} have been proposed to take advantage of game engine to construct large-scale synthetic re-ID datasets, which can be used to pre-train or fine-tune CNN network. In essence, it helps to provide more complete and better initialization parameters for potentially promoting the development of re-ID task. However, existing synthetic datasets have limited identities and lack of diversity, which leads to a large performance
degradation when transferring them to the wild or other scenes. Another challenge we observe is that, there exists serious scene shift between synthetic and real dataset. To be more specific, since synthetic dataset is large-attribute-range and diverse dataset, using all images may cause the side effect in real scenes. For instance, as shown in Fig.~\ref{fig1}, Market-1501 only contains scenes that recorded in summer vacation, DukeMTMC-reID is set in the blizzard scenes. Consequently, pre-training with all synthetic datasets can deteriorate performance on target domain during domain adaptation, which is not practical in real-world scenarios.

\begin{figure*}
\centerline{\includegraphics[width=\linewidth]{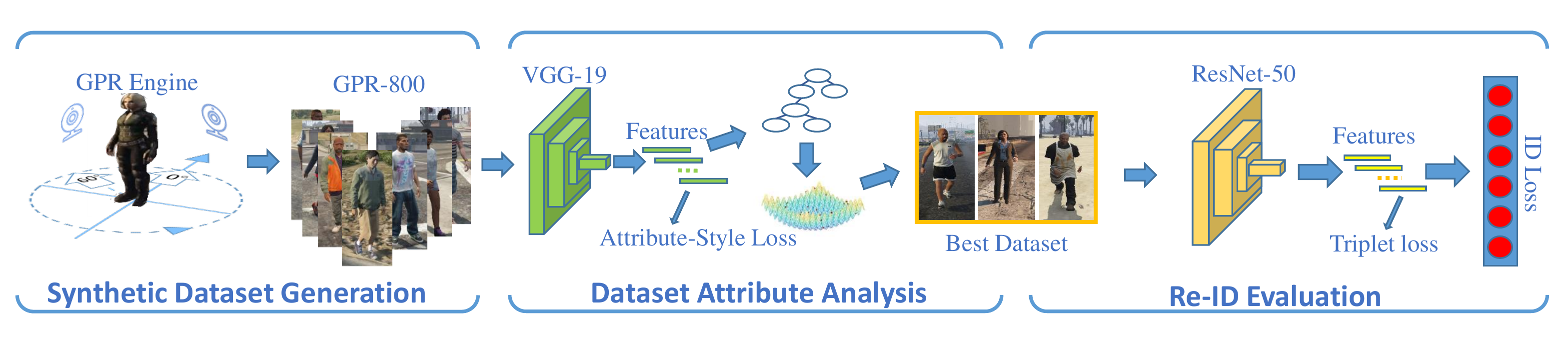}}
\caption{The procedure of our proposed end-to-end pipeline, which consists of 1) synthetic dataset generation, 2) dataset attribute analysis, and 3) re-ID evaluation period. Firstly, we employ GPR engine to generate a large-scale synthetic dataset named GPR-800. Based on it, we then adopt attribute-style loss with VGG-19 network to perform dataset attribute analysis, consequently, a more suitable dataset (reliable dataset) is constructed with prior knowledge of target domain. During the re-ID Evaluation stage, both triplet loss and ID loss are deployed to learn a discriminative re-ID model.}
\label{fig6}
\end{figure*}

To remedy the above problems, we start from two aspects. On the one hand, we introduce a large-scale synthetic data engine GPR-X, our virtual humans are carefully designed with an electronic game ``Grand Theft Auto V". Note that the salient characteristic of this engine is ``controllable". Based on it, we construct a large-scale and diverse synthetic person re-ID dataset named GPR-800. Compared with existing datasets, our GPR-800 has several advantages: 1) free colloection and annotation; 2) larger data volume; 3) more diversified scenes and 4) high resolution. More detailed information about GPR-800 is illustrated in Table~\ref{tabl}.

\begin{table}[]
  \centering
  \caption{Comparison of real-world and synthetic Re-ID datasets. ``View" denotes whether the dataset has viewpoint labels.}
  \small
  \setlength{\tabcolsep}{0.8mm}{
    \begin{tabular}{c|c|c|c|c|c}
    \Xhline{0.8pt}
    \multicolumn{2}{c|}{dataset} & \#identity & \#box & \#cam & view \\
    \Xhline{0.8pt}
    \multirow{3}[2]{*}{Real} & Market-1501~\cite{zheng2015scalable} & 1,501 & 32,668 & 6     & N \\
          & CUHK03~\cite{li2014deepreid} & 1,467 & 14,096 & 2     & N \\
          & DukeMTMC-reID~\cite{ristani2016performance}  & 1,404 & 36,411 & 8     & N \\
    \Xhline{0.8pt}
    \multirow{4}[2]{*}{Synthetic} & SOMAset~\cite{barbosa2018looking} & 50    & 100,000 & 250   & N \\
          & SyRI~\cite{bak2018domain}  & 100   & 1,680,000 & --    & N \\
          & PersonX~\cite{sun2019dissecting} & 1,266 & 273,456 & 6     & Y \\
          & \textbf{GPR-800}   & \textbf{800}   & \textbf{4,838,400} & \textbf{12}    & \textbf{Y} \\
    \Xhline{0.8pt}
    \end{tabular}}%
  \label{tabl}%
\end{table}%

On the other hand, in the attempt to reveal the influence of attributes of dataset on re-ID accuracy, we quantitatively analyze the influence of different factors (e.g. background, weather, illumination and viewpoint) on a person re-ID system with prior knowledge of target domain. To our knowledge there is no work in the existing literatures that comprehensively study the impacts of dataset attributes on re-ID system. So a natural question then come to our attention: \emph{how does these attributes influence the retrieval performance? Which one is most critical to our re-ID system?} To answer these questions, we perform rigorous quantification on pedestrian images regarding different dataset attributes. Both the control group and experimental group are designed, so as to obtain convincing scientific conclusions.

For the sake of demonstrating the potentiality of attribute analysis in recognizing pedestrian independently, we evaluate the performance based on the synthetic dataset in a discriminative re-ID system. To be more specific, the procedure of our proposed pipeline can be shown in Fig.~\ref{fig6}. Typically, the framework is not designed to achieve the-state-of-the-art performance in re-ID task, but to quantitatively analyze the influence of dataset attributes on re-ID system. To this end, we just select a simple but effective baseline for re-ID evaluation. The empirical results are consistent with our finding on the other CNN network.


As a consequence, this paper makes three contributions to the community.
\begin{itemize}
\item We introduce a large-scale and diverse synthetic dataset, which consists of 800 manually designed identities and editable visual variables.
\item Based on our synthetic dataset, we dissect a person re-ID system by quantitatively analyzing the influence of different attributes.
\item Comprehensive experiments conducted on benchmark datasets verify the effectiveness of proposed data-selecting strategy, helping us have a deeper
understanding of the fundamental problems in person re-ID.
\end{itemize}

The rest of the paper is organized as follows: After reviewing relevant previous works (Section 2), we describe the proposed GPR-X engine and GPR-800 synthetic dataset respectively (Section 3). Dataset attribute analysis is introduced in Section 4.1, Section 4.2 describes our baseline network for re-ID evaluation. Section 5 reports on an exhaustive set of experiments, illustrating the power of the attribute analysis strategy. Section 6 concludes with a summary and by sketching future work.
\begin{figure*}
\centerline{\includegraphics[width=\linewidth]{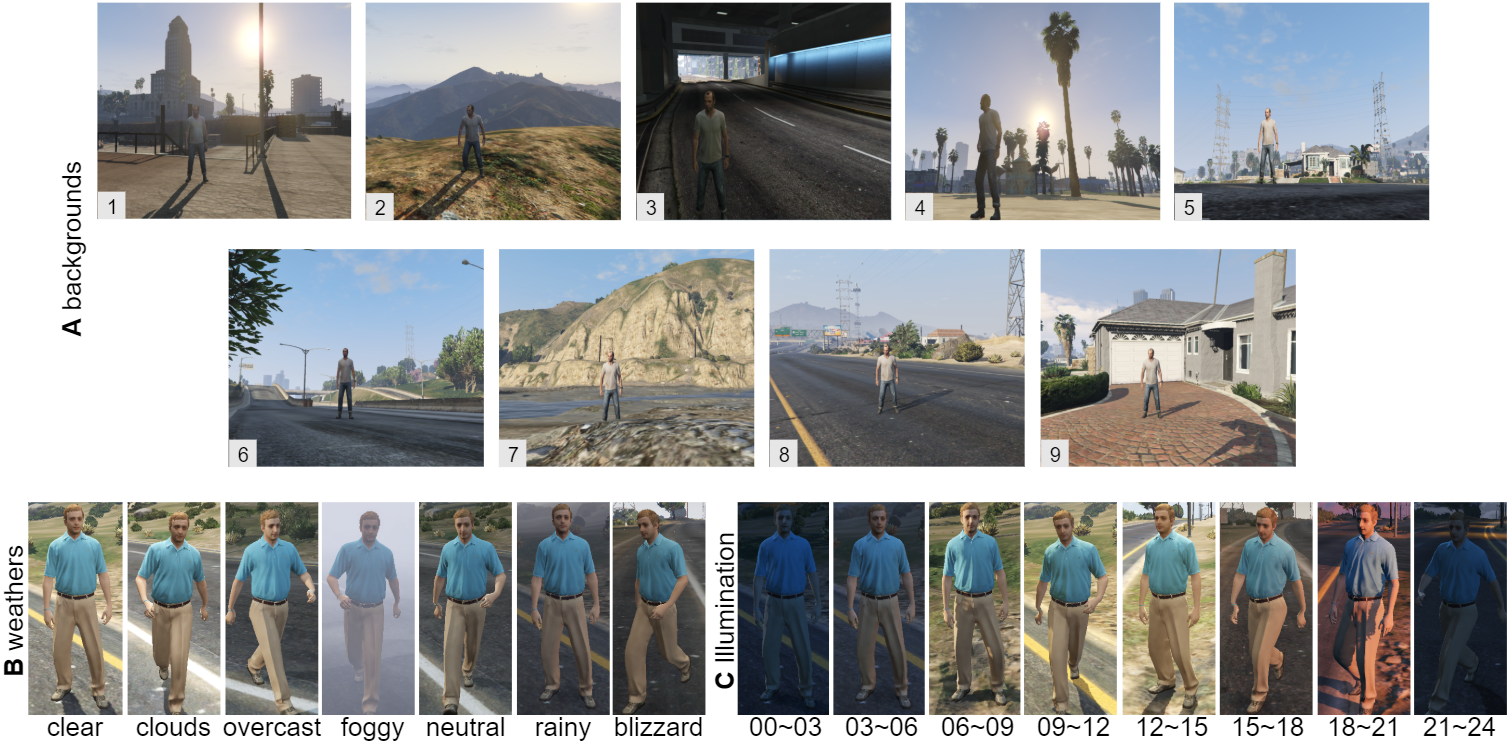}}
\caption{Illustration of GPR-800 dataset. (A): Background. In each background, a person can face toward a manually denoted direction. (B) and (C) denote the exemplars of different weather distribution and illumination distribution respectively.}
\label{fig2}
\end{figure*}

\section{Related Work}
\label{sec2}
It is known that manual labelling is generally time-consuming and labour-intensive for each new (target) domain. Transfer learning~\cite{xiang2019deep} sometimes, to some extent, works but fails to solve this problem fundamentally. More recently, leveraging synthetic data is a useful idea to alleviate the reliance on large-scale real datasets in person re-ID. Theoretically speaking, unlimited amount of labels can be made available by resorting to simulated data, which can greatly alleviate the problem of over-fitting caused by scarce real labeled data during training. In essence, the very recent re-ID approaches~\cite{barbosa2018looking,bak2018domain,sun2019dissecting} incorporate this idea to further boost re-ID performance in an unsupervised manner. For example, Barbosa et al.~\cite{barbosa2018looking} propose a synthetic instance dataset SOMAset, which is created by photorealistic human body generation software. Bak et al.~\cite{bak2018domain} construct a SyRI dataset by using 100 virtual humans illuminated with multiple HDR environment maps. In addition, Sun et al.~\cite{sun2019dissecting} introduce a large-scale synthetic data engine named PersonX. However, neither these synthetic datasets are intensively diversified, nor they are editable and extendable by the public. In comparison, our proposed GPR-800 synthetic dataset has configurable backgrounds, weathers, illuminations and diversified identities. More importantly, it can be extended not only for this study, but also lift the burden of constructing large-scale labeled datasets in this area, and free humans from heavy data annotations.

In recent years, visual attribute is an important factor in image retrieval due to its high-level semantic knowledge, which could greatly bridge the gap between low-level features and high-level human cognitions~\cite{gatys2016image,gatys2017controlling}. However,
the influence of background (context) regions of person images is mostly ignored by existing methods. Fortunately, in our experiment, we observed that, for a large person image database consisting of person images with different backgrounds, training with these datasets would greatly improve the robustness for person re-ID. Besides,
illumination~\cite{bak2018domain} is always a critical factor in person re-ID task, including viewpoints of pedestrian. However, current re-identification datasets lacks significant diversity in the number of lighting conditions or viewpoint angles, since they are usually limited to a relatively small number of cameras. Consequently, models trained on these special illuminations are thus biased to the limited illumination conditions seen during training, which fails to adapt the model to unseen illuminations. The same problem can be observed in terms of viewpoint~\cite{sun2019dissecting}.

To relieve this dilemma, in this paper, we firstly construct a large-scale synthetic dataset based on controllable synthetic data engine GPR-X, thus quantitatively analyze the influence of dataset attributes on re-ID accuracy. To the best of our knowledge, we are the first to conduct comprehensive experiments to quantitatively assess the influence of dataset attributes on person re-ID accuracy, which can help provide more meaningful guidance to construct a high-quality dataset for person re-ID task.

\begin{figure}
\centerline{\includegraphics[width=1.0\linewidth]{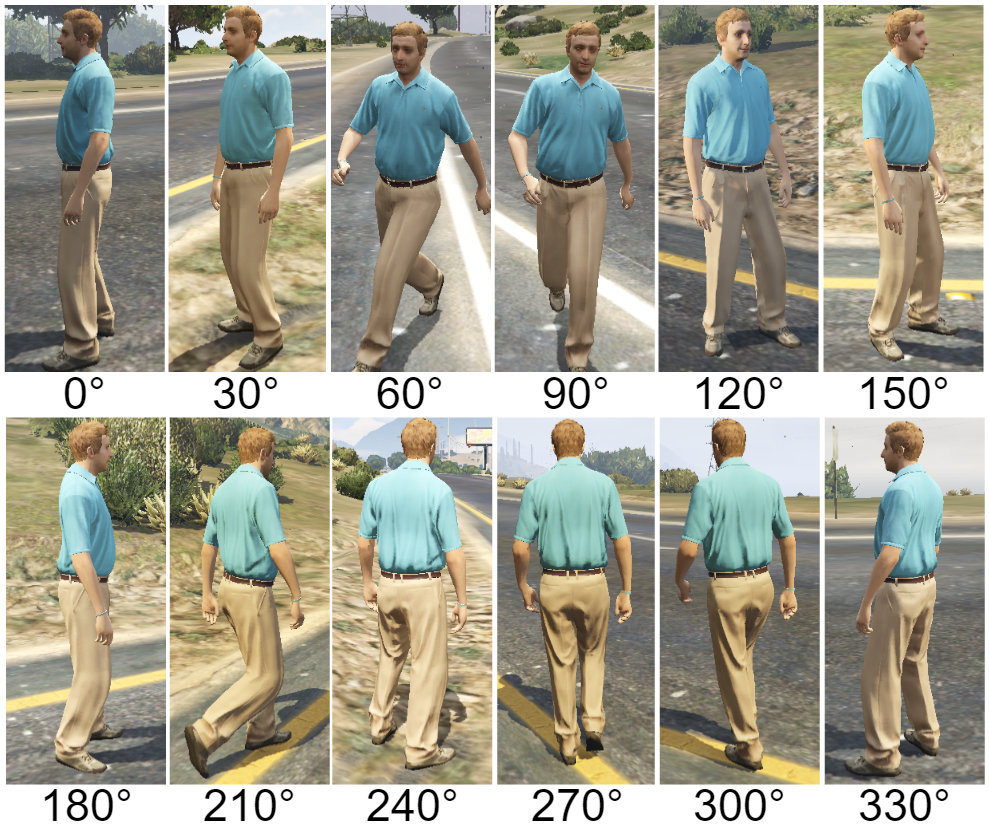}}
\caption{Image examples under a specified viewpoints, which is sampled every $30^{\circ}$ from $0^{\circ}$ to $330^{\circ}$ (12 different viewpoints in total).}
\label{fig3}
\end{figure}

\section{A Controllable Person Generation Engine}
\label{sec3}
\subsection{Description}
\textbf{Software.} The GPR-X engine is built on an electronic game Grand Theft Auto V, thus it is named as ``\textbf{G}TA5 \textbf{P}erson \textbf{R}e-identification" (``\textbf{GPR-X}" for short) engine. As a controllable system, it can satisfy various scene requirements. In GPR-X, the person models and scenes look realistic. More importantly, the values of visual variable, e.g., background, weather, illumination and viewpoint are designed to be editable, which allows GPR-X to be highly flexible and extendable.

\textbf{Identities.} GPR-X has 800 hand-crafted identities including different skin colors, body forms (e.g., height and weight) and hair styles, etc. To ensure diversity, the clothes of these identities include jeans, pants, shorts, slacks, skirts and different kinds of clothes, etc. In particular, some of these identities have a backpack, shoulder bag, glasses or hat. The motion of these characters can be walking, running, standing, or even having a dialogue etc. Fig.~\ref{fig1} (bottom) presents some examples of the character phototypes with customizable body parts and clothing. When not specified, all images are captured with resolution of 200 $\times$ 470.

\subsection{Visual Factors in GPR-X}
GPR-X is featured by editable environmental factors such as background, weather, illumination and viewpoint. More detailed information of these factors are described below.

\textbf{Backgound.}
Currently GPR-X has 9 different backgrounds, as shown in Fig.~\ref{fig2}(A). In each background, a person moves freely in arbitrary directions, exhibiting arbitrary viewpoints relative to the camera. In Fig.~\ref{fig2}(A), backgrounds (\emph{\#1}, \emph{\#4} and \emph{\#6}) depict different urban street scenes. Notably, adopting
more types of scenes close to target domain seems to have a positive influence on the re-ID performance.

\textbf{Weather.}
In our large-scale synthetic data engine GPR-X, there are 7 different types of weather, e.g., clear, clouds, overcast, foggy, neutral, rainy and blizzard. Parameters like degree and intensity can be modified for each weather type. By editing the values of these terms, various kinds of weather can be created. Some exemplars of synthetic scenes in different weather distribution from the proposed GPR-800 dataset are depicted in Fig.~\ref{fig2}(B).

\textbf{Illumination.}
Illumination can be obtained by in different time in whole day. For one specific pedestrian image, we also provide its capturing time in 24 hours. In this paper, we introduce a new synthetic dataset that contains 8 illumination conditions. The exemplars of different time distribution from the proposed GPR-800 dataset is illustrated in Fig.~\ref{fig2}(C), e.g., “09$\sim$12” denotes the time period during “9:00$\sim$12:00” in 24 hours a day.


\textbf{Viewpoint.}
Fig.~\ref{fig3} presents image examples under specified viewpoints following a uniform distribution. Those images are sampled during normal talking or walking. Specifically, a person image is sampled every $30^{\circ}$ from $0^{\circ}$ to $330^{\circ}$ (12 different viewpoints in total). Each view has 1 image, so each person has 12 images. The entire GPR-X engine has 800 (identities) $\times$ 9 (backgrounds) $\times$ 7 (weathers) $\times$ 8 (illuminations) $\times$ 12 (viewpoints) = 4,838,400 images.

The above discussions indicate that GPR-X engine has strictly controlled environment variables and reasonably sensitive to environmental changes. We believe GPR-X will be a useful tool for the community and encourage the development of robust algorithms and scientific analysis.

\section{Attribute Analysis}
\label{sec4}
It is known that attribute analysis on synthetic dataset can provide more meaningful guidance to construct a high-quality dataset for person re-ID task. Strictly speaking, finding that certain attributes is more important for learning models to identify pedestrians. For example, by discovering viewpoints that are effective for re-ID accuracy, our research can potentially benefit the practical usage of re-ID system. On the other hand, since GPR-800 is large-weather-range and diverse dataset, it is noteworthy that using all images to pre-train a CNN model will undoubtedly increase computational complexity, sometimes even causing the side effects in domain adaptation. In this section, we will quantitatively investigate the influences of different attributes on re-ID model learning that can potentially address this problem.

\begin{figure*}[t]
\begin{minipage}[t]{0.25\linewidth}
\centering
\includegraphics[width=1.9in]{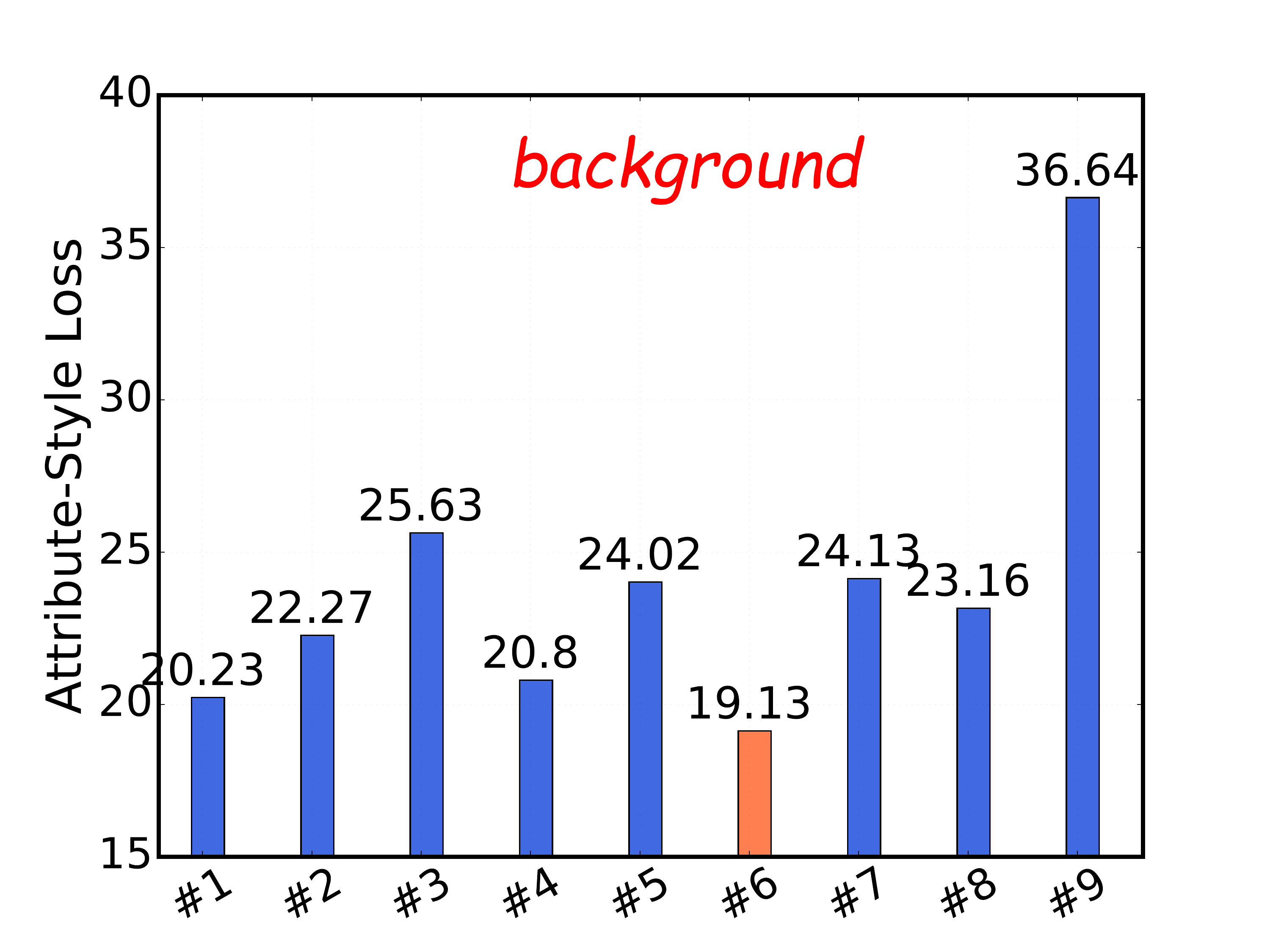}
\centerline{(a)}
\label{figa1}
\end{minipage}%
\begin{minipage}[t]{0.25\linewidth}
\centering
\includegraphics[width=1.9in]{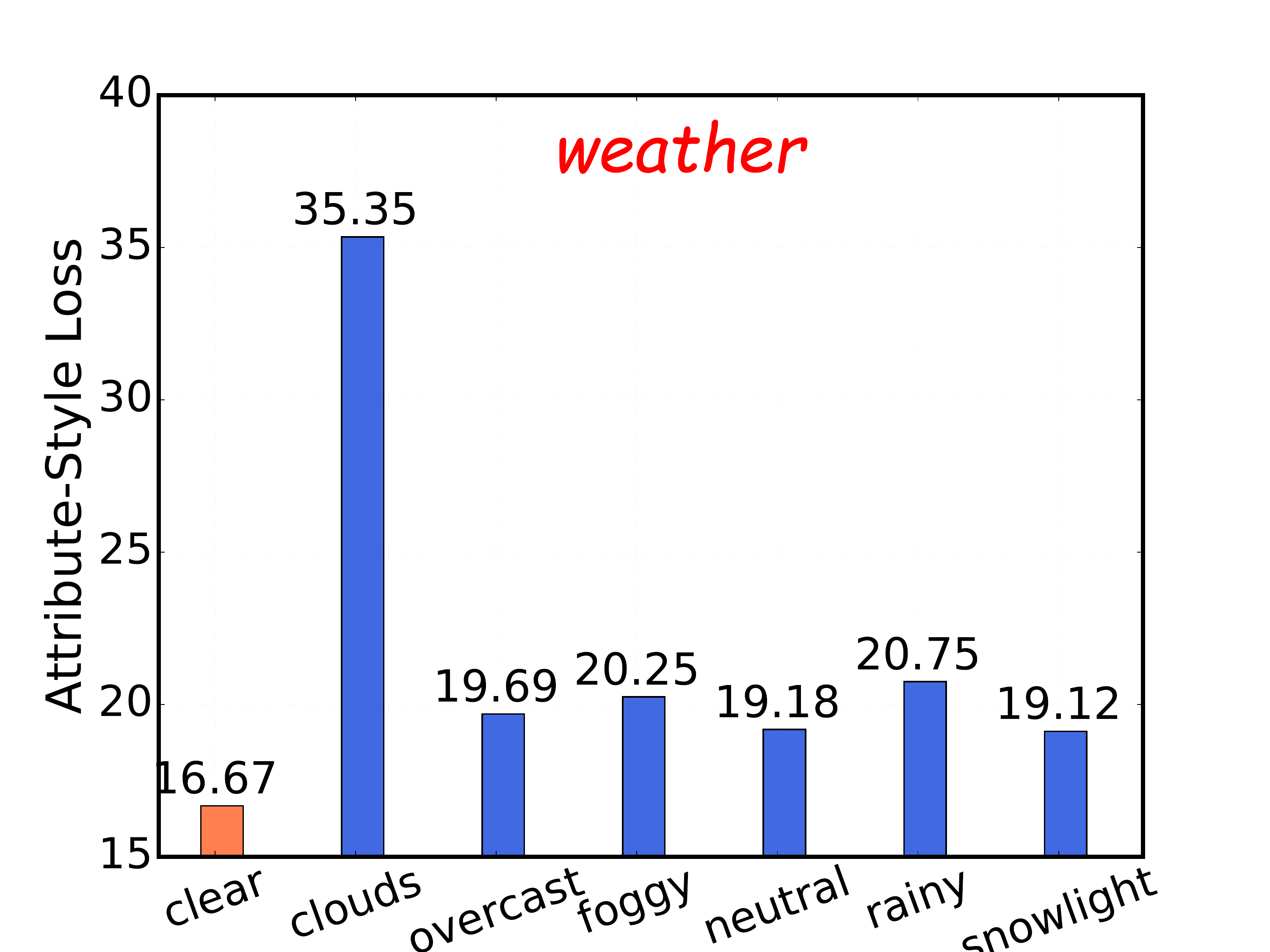}
\centerline{(b)}
\label{figa2}
\end{minipage}
\begin{minipage}[t]{0.25\linewidth}
\centering
\includegraphics[width=1.9in]{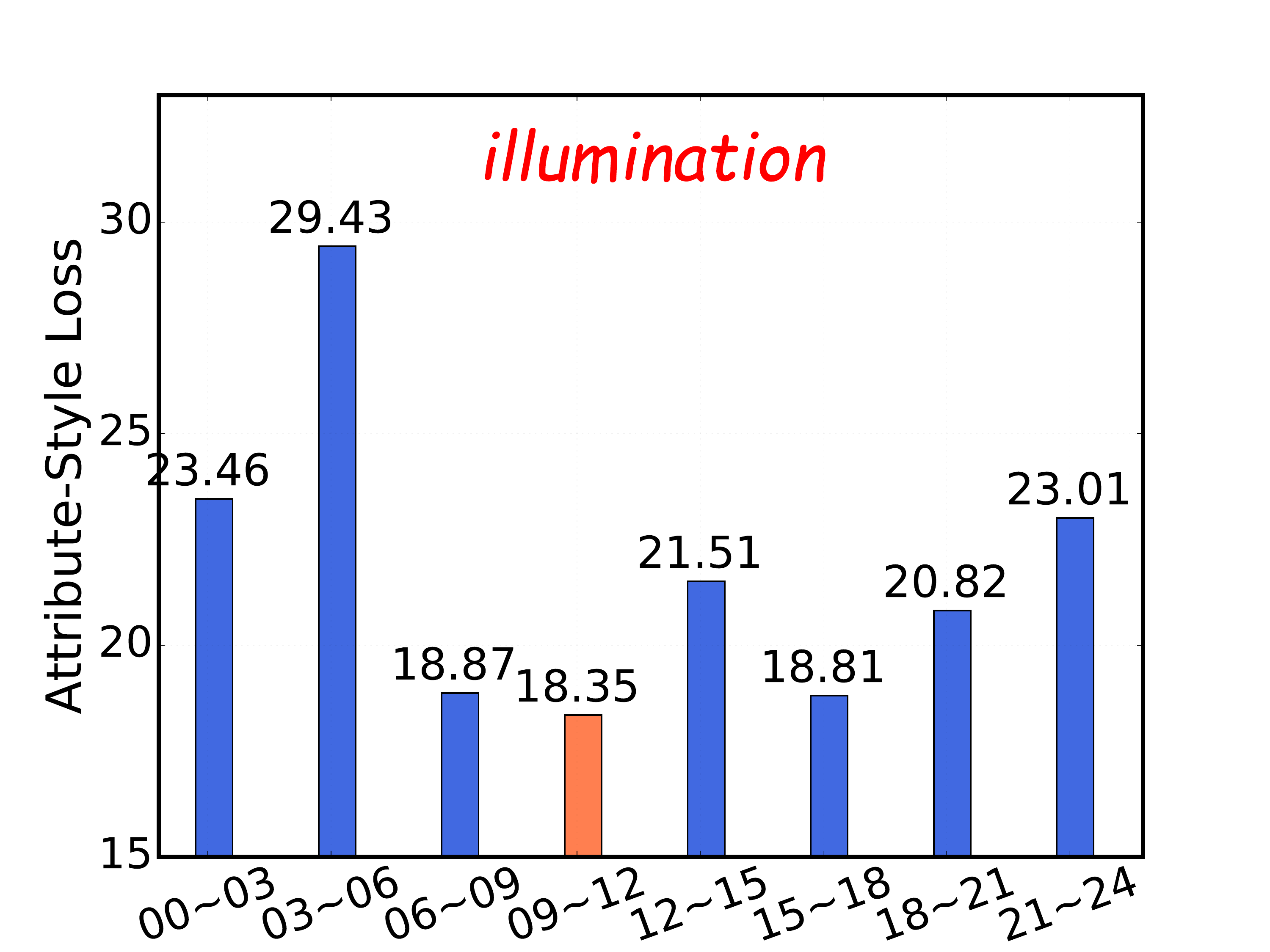}
\centerline{(c)}
\label{figa3}
\end{minipage}%
\begin{minipage}[t]{0.25\linewidth}
\centering
\includegraphics[width=1.9in]{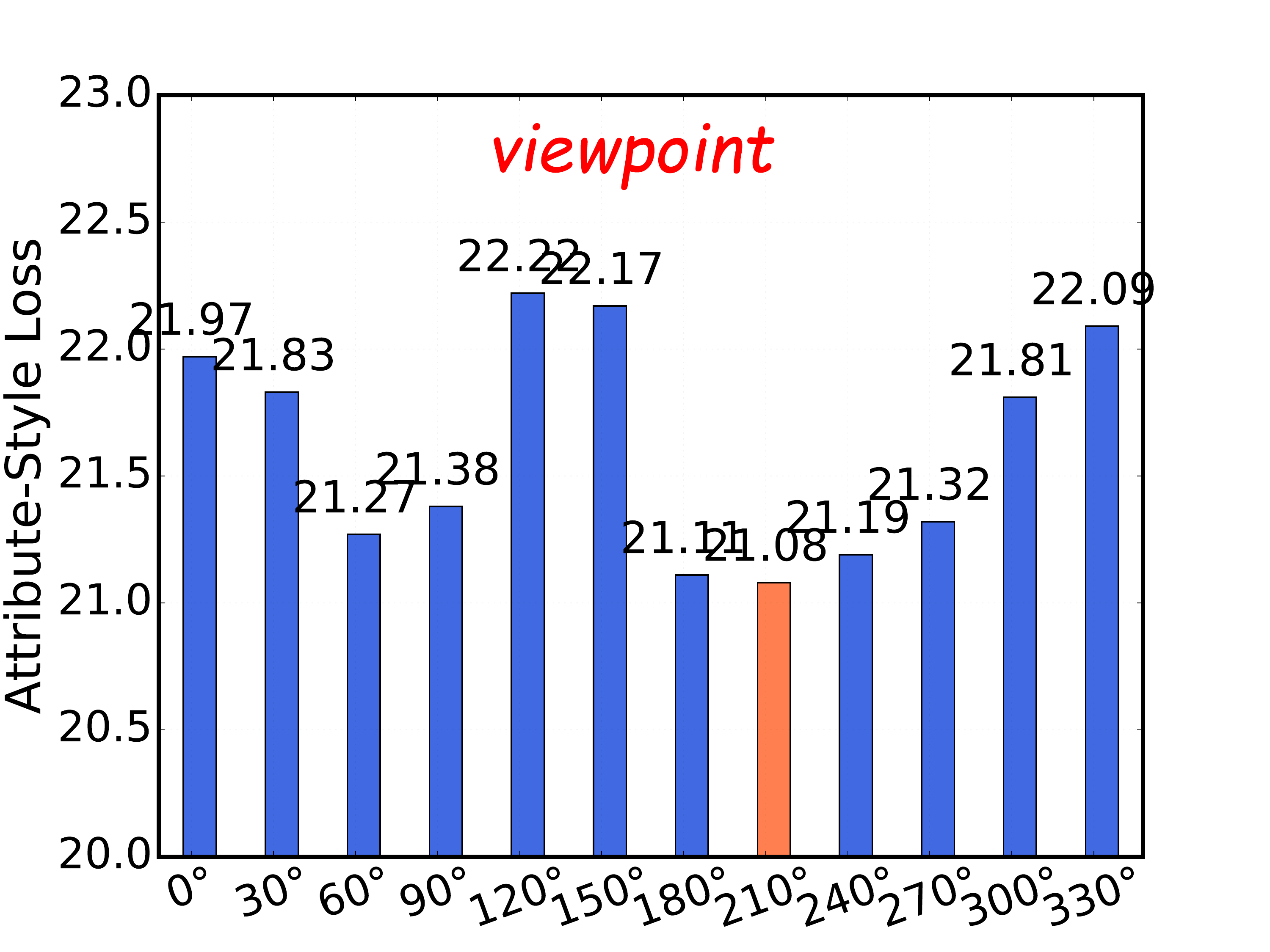}
\centerline{(d)}
\label{figa4}
\end{minipage}%
 \caption{Attribute analysis with style representation on VGG-19 when trained on synthetic GPR-800, while tested on \textbf{Market-1501}. It can be easily observed that the most critical factor in each datasets corresponds with items which have minimum loss in each attribute. \textcolor[rgb]{1.00,0.50,0.00}{\textbf{Orange}} in the bar chart indicates the most important factor in attribute analysis when performing re-ID task GPR-800 $\rightarrow$ Market-1501.}
\label{fig4}
\end{figure*}

\begin{figure*}[t]
\begin{minipage}[t]{0.25\linewidth}
\centering
\includegraphics[width=1.9in]{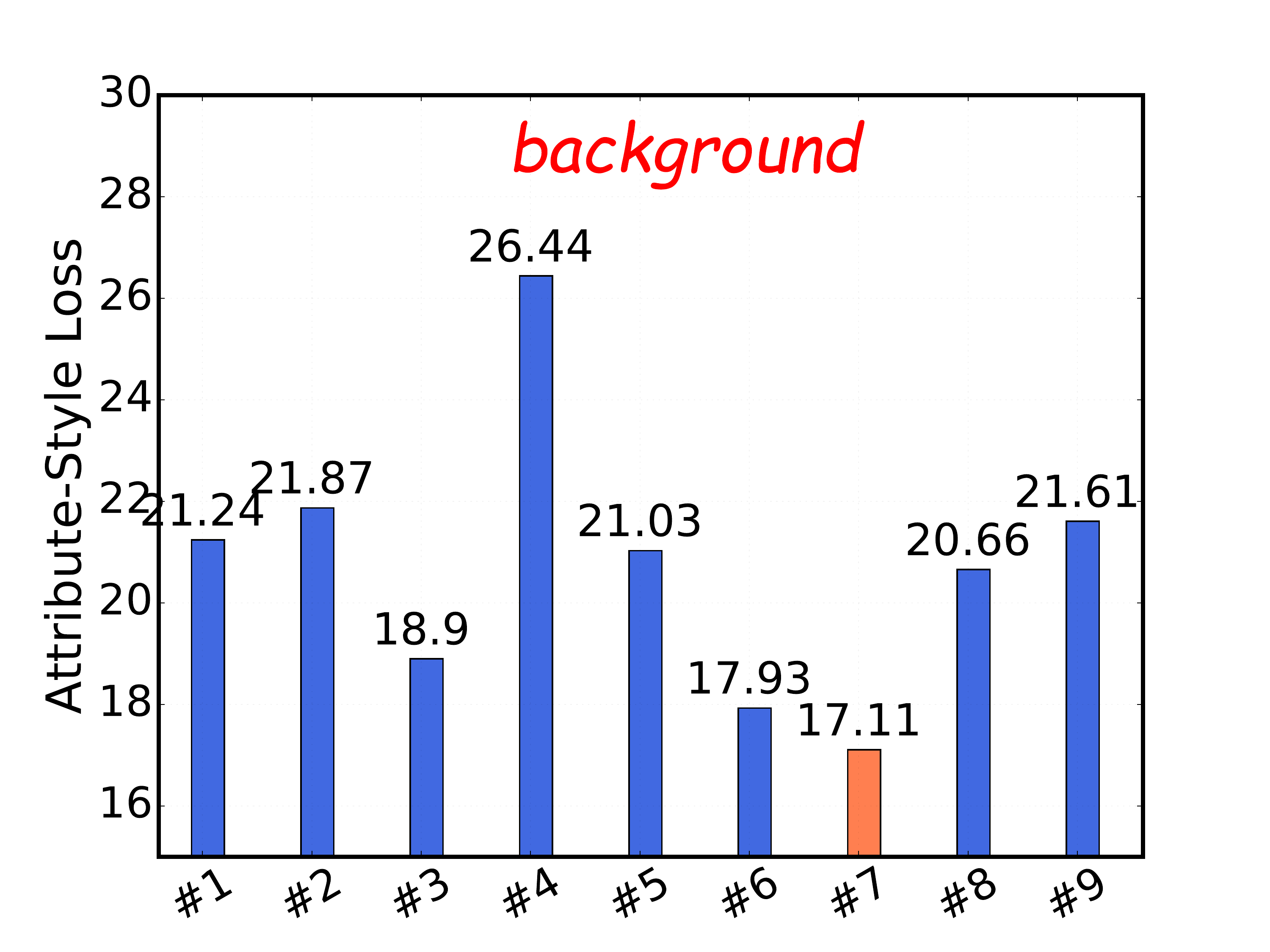}
\centerline{(a)}
\label{figb1}
\end{minipage}%
\begin{minipage}[t]{0.25\linewidth}
\centering
\includegraphics[width=1.9in]{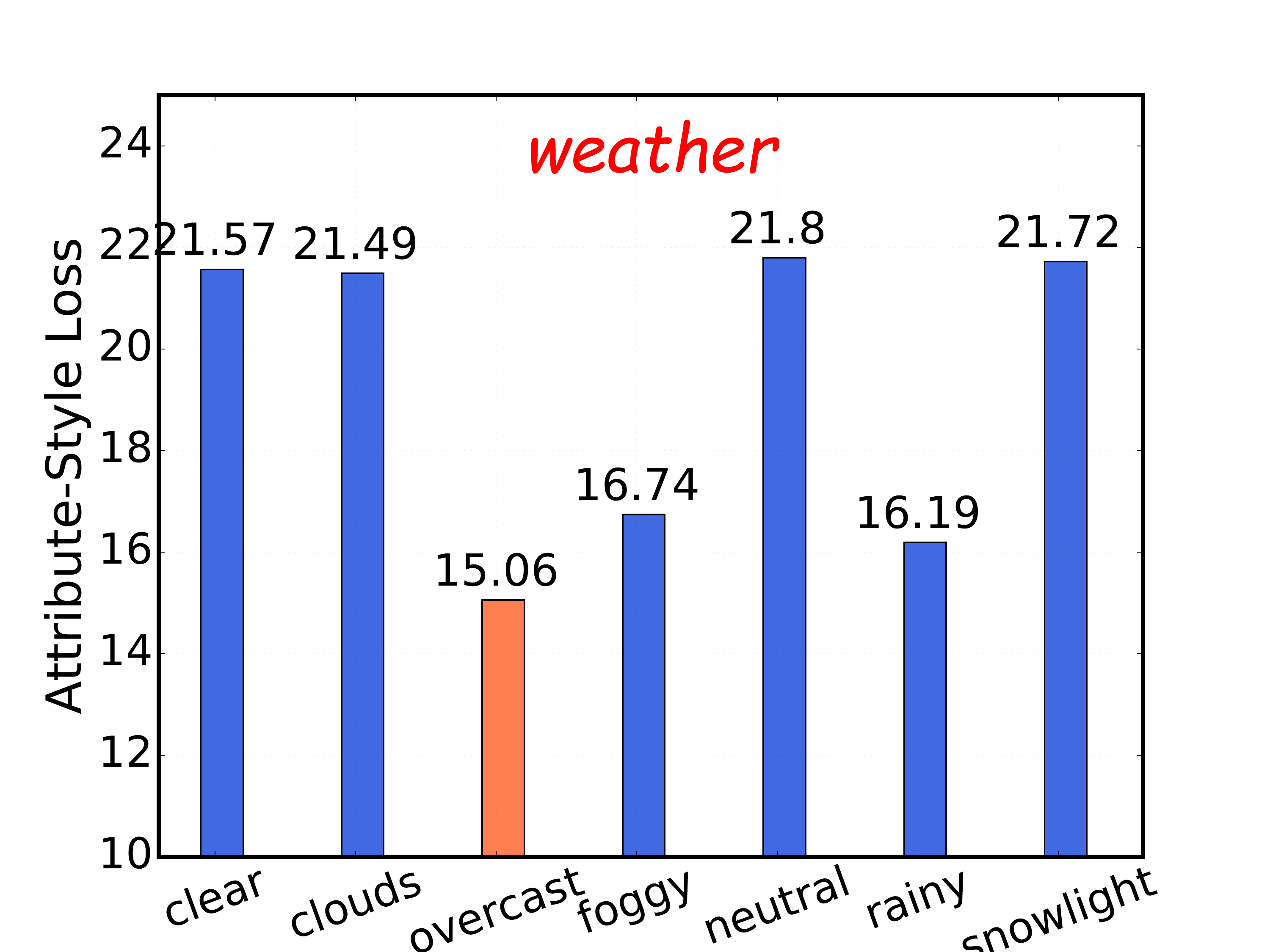}
\centerline{(b)}
\label{figb2}
\end{minipage}
\begin{minipage}[t]{0.25\linewidth}
\centering
\includegraphics[width=1.9in]{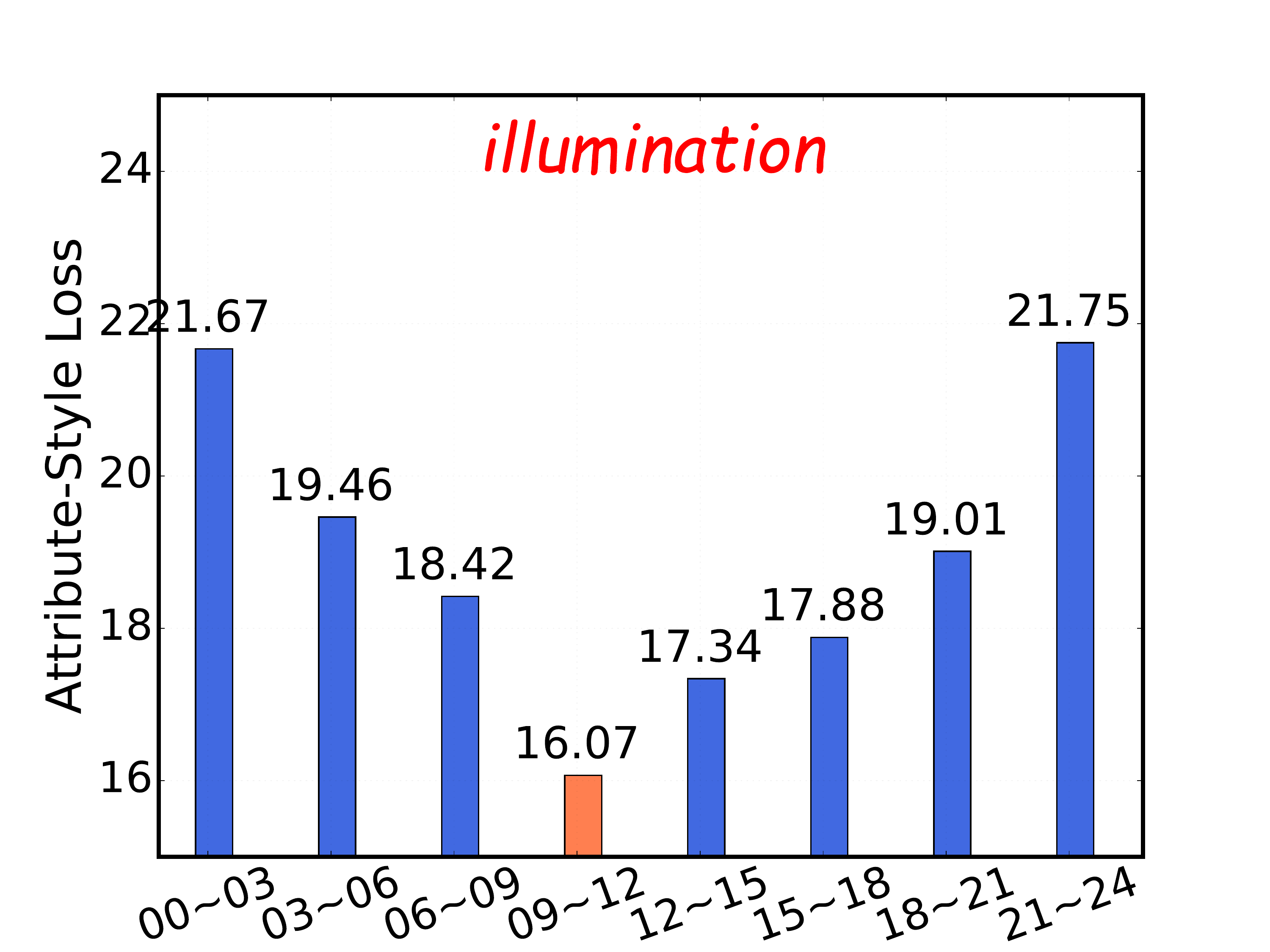}
\centerline{(c)}
\label{figb3}
\end{minipage}%
\begin{minipage}[t]{0.25\linewidth}
\centering
\includegraphics[width=1.9in]{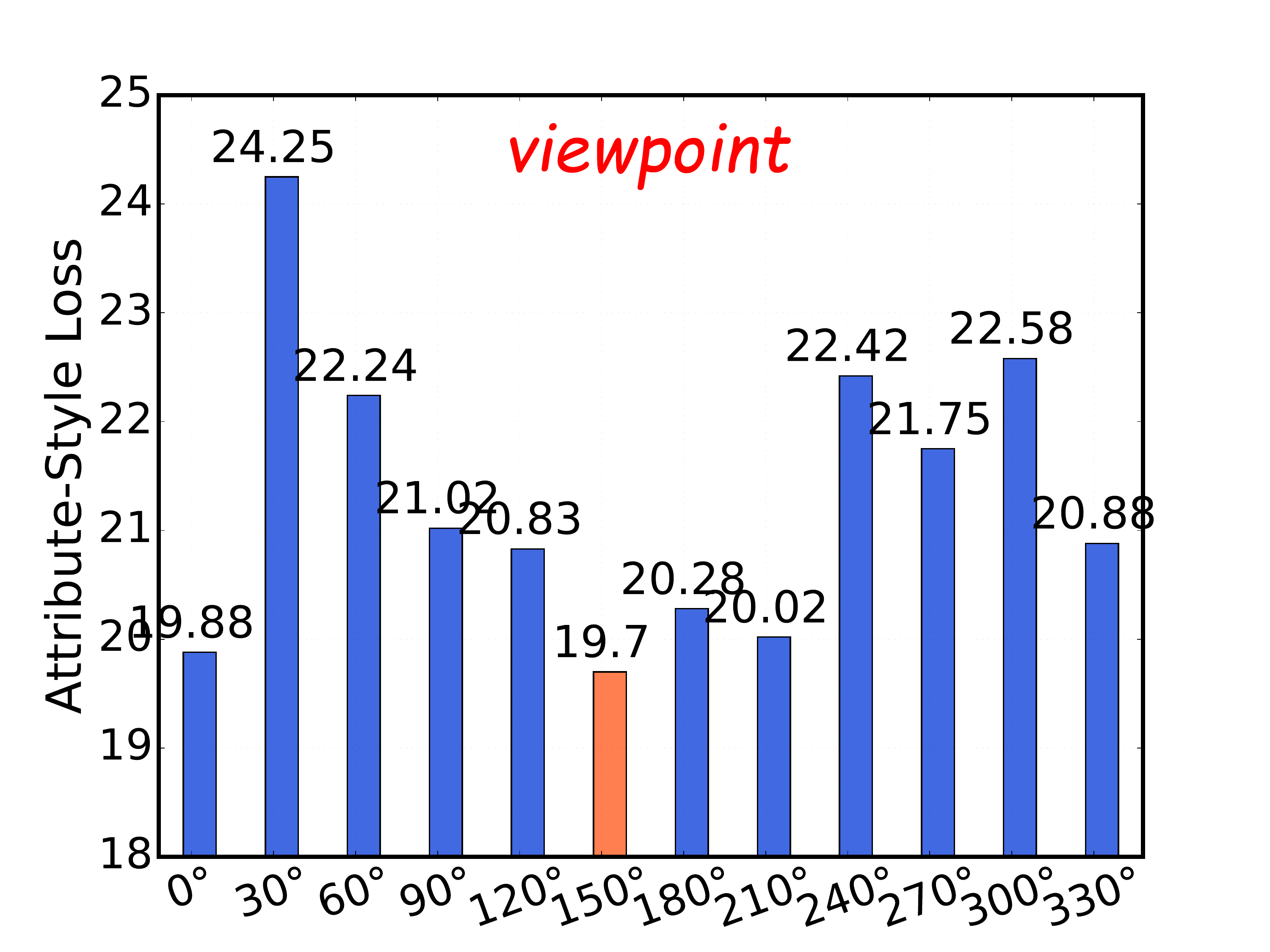}
\centerline{(d)}
\label{figb4}
\end{minipage}%
 \caption{Attribute analysis with style representation on VGG-19 when trained on synthetic GPR-800, while tested on \textbf{DukeMTMC-reID}. It can be easily observed that the most critical factor in each datasets corresponds with items which have minimum loss in each attribute. \textcolor[rgb]{1.00,0.50,0.00}{\textbf{Orange}} in the bar chart indicates the most important factor in attribute analysis when performing re-ID task GPR-800 $\rightarrow$ DukeMTMC-reID.}
\label{fig5}
\end{figure*}


\subsection{Attribute Analysis with Style Representation}
\label{sec4.1}
Style representation~\cite{gatys2016image,gatys2017controlling} has achieved remarkable results in the area of computer vision, which computes the correlations between the different filter responses. To obtain a representation of the style of an input image, we use a feature space designed to capture texture information, which is built on top of the CNN responses in each layer of the network. It consists of the correlations between the different filter responses, where the expectation is taken over the spatial extent of the feature maps. These feature correlations can be written as,

\begin{equation}
\label{eq1}
G_{i j}^{l}=\sum_{k} F_{i k}^{l} F_{j k}^{l}
\end{equation}
where $G^{l} \in \mathcal{R}^{N_{l} \times N_{l}}$ is Gram Matrix to define different feature correlations, $G_{i j}^{l}$ denote the inner product between the vectorised feature maps $i$ and $j$ in layer $l$, $F_{i j}^{l}$ is the activation of the $i^{th}$ filter at position $j$ in layer $l$.

Indeed, we can visualise the information captured by these style feature spaces built on different layers of the network. To generate a texture that matches the style of a given image, we use gradient descent from a white noise image to find another image that matches the style representation of original image. This is done by minimising the mean-squared distance between the entries of the Gram Matrix from the original image and the Gram Matrix of the target real image. To this end, the contribution of that layer to the total loss is then written as,
\begin{equation}\label{eq2}
E_{l}=\frac{1}{4 N_{l}^{2} M_{l}^{2}} \sum_{i, j}\left(G_{i j}^{l}-A_{i j}^{l}\right)^{2}
\end{equation}
where $A^{l}$ and $G^{l}$ denote their respective style representation in lay $l$ respectively, $N_{l}$ indicates the number of distinct filters, $M_{l}$ is the product of height $\times$ the width of the feature map.
And the total attribute-style loss is written as,
\begin{equation}\label{eq3}
\mathcal{L}_{\text {style}}(\vec{a}, \vec{x})=\sum_{l=0}^{L} w_{l} E_{l}
\end{equation}
where $w_{l}$ is weighting factors of the contribution of each layer to the total loss, $\vec{a}$ and $\vec{x}$ denote be the original image and the image that is in target domain.

After obtaining the attribute-style loss between the GPR-800 dataset and the target dataset, we firstly conduct a simple statistical analysis, as depicted in Fig.~\ref{fig4} and Fig.~\ref{fig5}. It can be easily observed that the datasets with different attributes have a distinct distribution in a specific task. For example, when performing adaptation from GPR-800 $\rightarrow$ Market-1501, \emph{background ({\#1}, \#4 and \#6)} is more sensitive since it has relatively smaller attribute-style loss, as shown in Fig.~\ref{fig4}(a). In the Subsection 5.2, we will quantitatively verify the effectiveness of attribute analysis strategy in dissecting person re-ID task.

\subsection{re-ID Evaluation}
\label{sec4.2}
Many existing re-ID approaches~\cite{zhong2018generalizing} are based on a model pre-trained on ImageNet~\cite{deng2009imagenet}, and we follow the similar setting in~\cite{luo2019bag} to obtain a initializing model. In particular, the last fully connected layer is discarded and two additional FC layer are added. To be more specific, the first FC layer has 2,048 dimensions, which is training with cross-entropy loss~\cite{zhang2018generalized} by casting training process as a classification problem, while the second FC is $P_{s}$ dimensional, where $P_{s}$ is the number of identity in source dataset, batch-hard triplet loss~\cite{hermans2017defense} is employed with second FC layer by treating the training process as a verification problem, this gives rise to the re-ID evaluation stage of our proposed pipeline (Fig.~\ref{fig6}).

\begin{table*}[htbp]
  \centering
  \caption{Ablation study on Market-1501. After adopted our attribute analysis with style representation, we can find the most critical factor when adapting our model to Marlet-1501. Specifically, \CheckmarkBold indicates using all conditions in each attribute. Measured by \%.}
  \small
  \setlength{\tabcolsep}{1.2mm}{
    \begin{tabular}{c|c|c|c|c|c|c|c|c}
    \Xhline{0.8pt}
    \#identity & \#box   & \#background & \#weather & \#illumination & \#viewpoint & mAP   & rank-1   & rank-5\\
    \Xhline{0.8pt}
    100   & 604,800 & \CheckmarkBold    & \CheckmarkBold     & \CheckmarkBold     & \CheckmarkBold     & 4.5   & 12.9 &   27.6 \\
    \hline
    400   & 2,419,200 & \CheckmarkBold     & \CheckmarkBold     & \CheckmarkBold     & \CheckmarkBold     & 4.7   & 14.1   & 28.3 \\
    \hline
    700   & 4,233,600 & \CheckmarkBold     & \CheckmarkBold     & \CheckmarkBold     & \CheckmarkBold     & 11.4  & 31.3   & 49.5 \\
    \hline
    100    & 67,200 & \#6     & \CheckmarkBold     & \CheckmarkBold     & \CheckmarkBold     & 7.3   & 21.2   & 36.4 \\
    \hline
    100    & 86,400 & \CheckmarkBold     & clear     & \CheckmarkBold     & \CheckmarkBold     & 8.8   & 24.1   & 38.9 \\
    \hline
    100   & 75,600 & \CheckmarkBold     & \CheckmarkBold     & 09 $\sim$ 12     & \CheckmarkBold     & 9.7   & 26.3   & 42.7 \\
    \hline
    100   & 302,400 & \CheckmarkBold     & \CheckmarkBold     & \CheckmarkBold     & $60^{\circ}$, $90^{\circ}$,$180^{\circ}$,$210^{\circ}$,$240^{\circ}$, $270^{\circ}$     & 10.2   & 28.5   & 45.0 \\
    \hline
    800   & 115,200 & \#1,\#4,\#6     & clear,neutral     & 06 $\sim$ 18     &  $60^{\circ}$, $90^{\circ}$,$180^{\circ}$,$210^{\circ}$,$240^{\circ}$, $270^{\circ}$      & \textbf{14.3}  & \textbf{34.2}   & \textbf{58.5} \\
    \Xhline{0.8pt}
    \end{tabular}}%
  \label{tab3}%
\end{table*}%

\begin{table*}[htbp]
  \centering
  \caption{Ablation study on DukeMTMC-reID. After adopted our attribute analysis with style representation, we can find the most critical factor when adapting our model to Marlet-1501. Specifically, \CheckmarkBold indicates using all conditions in each attribute. Measured by \%. }
  \small
  \setlength{\tabcolsep}{1.2mm}{
    \begin{tabular}{c|c|c|c|c|c|c|c|c}
    \Xhline{0.8pt}
    \#identity & \#box   & \#background & \#weather & \#illumination & \#viewpoint & mAP   & rank-1   & rank-5\\
    \Xhline{0.8pt}
    100   & 604,800 & \CheckmarkBold     & \CheckmarkBold     & \CheckmarkBold     & \CheckmarkBold     & 3.8   & 11.9 &   20.4 \\
    \hline
    400   & 2,419,200 & \CheckmarkBold     & \CheckmarkBold     & \CheckmarkBold     & \CheckmarkBold     & 9.3   & 24.1   & 37.3 \\
    \hline
    700   & 4,233,600 & \CheckmarkBold     & \CheckmarkBold     & \CheckmarkBold     & \CheckmarkBold     & 14.8  & 33.5   & 47.2 \\
    \hline
    100    & 67,200 & \#7     & \CheckmarkBold     & \CheckmarkBold     & \CheckmarkBold     & 5.9   & 13.7   & 26.1 \\
    \hline
    100    & 86,400 & \CheckmarkBold     & overcast     & \CheckmarkBold     & \CheckmarkBold     & 5.7   & 13.6   & 25.8 \\
    \hline
    100   & 75,600 & \CheckmarkBold     & \CheckmarkBold     & 09 $\sim$ 12     & \CheckmarkBold     & 11.3   & 24.6   & 38.6 \\
    \hline
    100   & 302,400 & \CheckmarkBold     & \CheckmarkBold     & \CheckmarkBold     & $0^{\circ}$,$90^{\circ}$,$120^{\circ}$,$150^{\circ}$,$180^{\circ}$, $210^{\circ}$     & 12.6   & 28.3   & 44.8 \\
    \hline
    800   & 115,200 & \#3,\#6,\#7     & overcast,rainy     & 06 $\sim$ 18     &  $0^{\circ}$,$90^{\circ}$,$120^{\circ}$,$150^{\circ}$,$180^{\circ}$, $210^{\circ}$       & \textbf{14.2}  & \textbf{31.7}   & \textbf{46.6} \\
    \Xhline{0.8pt}
    \end{tabular}}%
  \label{tab4}%
\end{table*}%

\section{Experiments and Evaluation}\
\label{sec5}
In this paper, we evaluate our method on two large-scale benchmark datasets, Market-1501~\cite{zheng2015scalable} and DukeMTMC-reID~\cite{ristani2016performance,zheng2017unlabeled}.

\textbf{Market-1501} This dataset has 32,668 person images of 1,501 identities. There are 6 cameras views in the summer campus. As official setting, 751 ids are used for training and the rest 750 ids are used for testing. The query contains 3,368 images.

\textbf{DukeMTMC-reID} This dataset is constructed from the multi-camera tracking dataset DukeMTMC, it contains 1,812 identities. 702 identities are used as the training set and the remaining 1,110 identities as the testing set. It contains 36,411 images in total, 2,228 images are used as queries.

\textbf{Protocols} In the experiment, we follow the standard evaluation protocol used in~\cite{zhong2017re} and adopt mean Average Precision (mAP) and Cumulative Marching Characteristics (CMC) at rank-1, rank-5 for performance evaluation on all the candidate datasets.

\subsection{Experiments Setting}
In experiment, we adopted PyTorch~\cite{paszke2017automatic} to implement and train our re-ID learning network. The training procedure is standard and requires no bells and whistles. For attribute analysis, attribute-style loss presented in Fig.~\ref{fig4} and Fig.~\ref{fig5} were generated on the basis of the VGG network~\cite{simonyan2014very}, and we empirically set $w_{l}$ = 0.2 in Eq.~\ref{eq3}. During evaluation stage, our network is modified based on the ResNet-50~\cite{he2016deep}, following the baseline training strategy introduced in~\cite{luo2019bag} on Tesla P100 GPU. In particular, we keep the aspect ratio of input images and resize them to 128$\times$64. Note that we only employ random erasing to training set for data augmentation, and choose SGD as the optimizer with momentum 0.9. The weight decay factor for L2 regularization is set to 0.0005.


\subsection{Evaluation}
\label{sec5.1}
We evaluate the impacts of different attributes on a basic person re-ID system. Note that the experimental groups are used to assess the impact of different attributes in the training set. For a clearer understanding, we will demonstrate more comprehensive results with both figures and tables.

\textbf{Experiment design.} We train a re-ID model on the original synthetic dataset comprised of 4 different attributes, e.g., background, weather, illumination
and viewpoint. We tested on one of attributes while keep the rest fixed. Before performing re-ID evaluation, we first employ style representation to select some important factors. For instance, when tested on Market-1501, we calculate the attribute-style loss $\mathcal{L}_{\text {style}}$ with Eq.~\ref{eq3}. Each attribute with relatively smaller attribute-style loss can be regarded as key factors. As shown in Fig.~\ref{fig4}, \emph{background (\#6)}, \emph{weather (clear)}, \emph{illumination (09$\sim$12)} and \emph{viewpoint ($210^{\circ}$)} tend to have relatively smaller attribute-style loss. We argue that, it is probably because that this specific attributes set is more close to real market-1501 scene. Consequently, it undoubtedly leads to much better performance when adapted to Market-1501.

\subsection{Effectiveness of Attribute Analysis}
\label{sec5.2}
In the section, we further explore the effectiveness of attribute analysis. Table~\ref{tab3} and Table~\ref{tab4} present the results obtained by the above attribute analysis strategy. There are several observations which can be made as follows.

First, we can easily observe that using more IDs will noticeably improve the re-ID performance. For example, as presented in Table~\ref{tab3} and Table~\ref{tab4}, we can only achieve a performance of 4.5\% and 3.8\% in mAP accuracy when tested on Market-1501 and DukeMTMC-reID, respectively. Moreover, adding IDs to 700 as supervised information notably improves the re-ID accuracy, leading to \textbf{+6.9\%} and \textbf{+11.0\%} in mAP accuracy.

Second, with constraint of ID number equals to 100, using \emph{background (\#6)} and \emph{background (\#7)} bring about +2.8\% and +2.1\% more improvement than using all backgrounds;
using \emph{weather (clear)} and \emph{weather (overcast)} as constriant can take additional improvement of +4.3\% and +1.9\% in mAP accuracy. Same conclusion can also be drew by taking ``illumination" into consideration. Furthermore, we can achieve a significant improvement of \textbf{+15.6\%} and \textbf{+16.4\%} in rank-1 accuracy by selecting some critical viewpoints when tested on Market-1501 and DukeMTMC-reID dataset. In other words, the major observation is that the retrieval accuracy will be negatively affected and there will be a non-trivial performance drop without these attribute constraints, which verifies the effectiveness of our proposed attribute analysis strategy.

To this end, by taking background, weather, illumination and viewpoint into account, we can further boost the performance to 14.3\% and 14.2\% when performing GPR-800 $\rightarrow$ Market-1501 and GPR-800 $\rightarrow$ DukeMTMC-reID, respectively. Compared with training only one attribute constraint, our multiple attribute constraint can achieve better performance with less labeled data for training, which can save lots of time cost for training and human labor for dataset labelling. This conclusion is very meaningful since it will provide guidance for us to construct a high-quality re-ID dataset. Furthermore, we can drastically improve our performance by enhancing the diversity of re-ID train-set in the further research. Nevertheless, we find it truly fascinating that a re-ID system, which is trained to perform one of the core computational tasks of image retrieval, automatically learns image representations that allow, at least to some extent, the separation of image attribute from content.

\subsection{Discussion}
\label{sec5.3}
The main purpose of this paper is to evaluate the effect of different attributes on re-ID system with a simple baseline, so high performance in domain adaptation is not our point. In essence, performance of re-ID model trained on synthetic dataset, while tested on Market-1501 or DukeMTMC-reID may not be so competitive when compared with the SOTA methods~\cite{zhong2019invariance,yu2019unsupervised, xiang2020unsupervised}, which is trained on real dataset, we argue that, it probably because there exists a huge gap between synthetic and real image distribution, so learning an invariant feature is extremely difficult when directly performing adaptation from synthetic to real domain.

\section{Discussion and Conclusion}
\label{sec6}
This paper makes a step from engineering new technologies to science new discoveries. We make two contributions to the community. First, we build a synthetic data engine GPR-X than can generate images under controllable cameras and environments. Second, based on GPR-X, we conduct comprehensive experiments to quantitatively assess the influence of dataset attributes on person re-ID accuracy. To be more specific, we can find most critical factors by style representation when given a specific target domain, which can greatly decrease the scale of training samples, and save time cost for training and human label expenses for labelling. In the future, we will further explore other style transferring methods to bridge the gap between synthetic and real dataset, and boost the performance for re-ID tasks .

\bibliographystyle{IEEEtran}
\bibliography{references}{}

\end{document}